\pdfoutput=1
\documentclass[11pt]{article}

\usepackage[preprint]{acl}

\usepackage{times}
\usepackage{latexsym}

\usepackage[T1]{fontenc}
\usepackage[utf8]{inputenc}

\usepackage{microtype}

\usepackage{inconsolata}

\usepackage{graphicx}

\usepackage{amsmath}
\usepackage{booktabs}
\usepackage{siunitx}
\usepackage{pgfplots}
\pgfplotsset{compat=1.18}
\usepackage{algorithm}
\usepackage{algpseudocode}
\usepackage{xcolor}
\usepackage{pgf-pie}
\usepackage{CJKutf8}
\usepackage{multirow}
\usepackage{soul}
\usepackage{tcolorbox}
\tcbuselibrary{listings}
\usepackage{subcaption}
\usepackage{tipa}
\usepackage{arabtex}
\usepackage{utf8}
\setcode{utf8}

\newcommand{\ja}[1]{\begin{CJK}{UTF8}{min}#1\end{CJK}}

\title{Languages Still Left Behind: Toward a Better Multilingual Machine Translation Benchmark}

\author{
    \textbf{Chihiro Taguchi\textsuperscript{1}},
    \textbf{Seng Mai\textsuperscript{2}},
    \textbf{Keita Kurabe\textsuperscript{3}},
    \textbf{Yusuke Sakai\textsuperscript{4}},
\\
    \textbf{Georgina Agyei\textsuperscript{1*}}
    \textbf{Soudabeh Eslami\textsuperscript{5*}},
    \textbf{David Chiang\textsuperscript{1}}
\\
\\
    \textsuperscript{1}University of Notre Dame,
    \textsuperscript{2}University of Washington,
    \textsuperscript{3}Tokyo University of Foreign Studies,
\\
    \textsuperscript{4}Nara Institute of Science and Technology,
    \textsuperscript{5}University of T{\"u}bingen
\\
    \small{
        \textbf{Correspondence:} \href{mailto:ctaguchi@nd.edu}{ctaguchi@nd.edu}
    }
}

\begin{document}

\maketitle
\def\thefootnote{*}\footnotetext{Equal contribution.}
\def\thefootnote{\arabic{footnote}}

\begin{abstract}
Multilingual machine translation (MT) benchmarks play a central role in evaluating the capabilities of modern MT systems.
Among them, the FLORES+ benchmark is widely used, offering English-to-many translation data for over 200 languages, curated with strict quality control protocols.
However, we study data in four languages (Asante Twi, Japanese, Jinghpaw, and South Azerbaijani) and uncover critical shortcomings in the benchmark's suitability for truly multilingual evaluation.
Human assessments reveal that many translations fall below the claimed 90\% quality standard, and the annotators report that source sentences are often too domain-specific and culturally biased toward the English-speaking world.
We further demonstrate that simple heuristics, such as copying named entities, can yield non-trivial BLEU scores, suggesting vulnerabilities in the evaluation protocol.
Notably, we show that MT models trained on high-quality, naturalistic data perform poorly on FLORES+ while achieving significant gains on our domain-relevant evaluation set.
Based on these findings, we advocate for multilingual MT benchmarks that use domain-general and culturally neutral source texts rely less on named entities, in order to better reflect real-world translation challenges.\footnote{
The code, the assessment results, and the new datasets used in this work are available at \url{https://github.com/ctaguchi/LSLB}.
}
\end{abstract}

\section{Introduction}

Recent advances in natural language processing have been increasingly multilingual in nature, and machine translation (MT) is no exception. To assess the multilingual capabilities of MT models, several evaluation benchmarks have been proposed, among which the FLORES+ benchmark \citep{guzman-etal-2019-flores, goyal-etal-2022-flores} has emerged as one of the most prominent.
FLORES+ provides a large-scale English-to-many translation dataset covering over 200 languages.
The creation process of the dataset adheres to strict quality control standards, including rigorous qualification requirements for both translators and reviewers, and the benchmark claims to maintain a translation quality score of over 90\%.

In this paper, we carry out a re-evaluation of the FLORES+ data in four languages (Asante Twi, Japanese, Jinghpaw, and South Azerbaijani).
Our study reveals significant concerns regarding the adequacy and fluency of the benchmark for truly multilingual MT evaluation.
We observe that the translation quality in the FLORES+ dataset often falls short of the reported 90\% threshold.
Feedback from evaluators further highlights that many of the original English source sentences are overly domain-specific and frequently contain jargon or culturally specific expressions that are difficult or impossible to translate naturally into the target languages.

Moreover, we identify structural issues in the benchmark itself.
Through a simple experiment, we show that models can achieve the average BLEU score of 0.29 merely by copying named entities (NEs) from the source sentence, calling into question the benchmark's robustness.
More critically, our experiments on Jinghpaw MT demonstrate that models trained on naturalistic, high-quality translation data underperform on the FLORES+ benchmark but significantly outperform on a separate, out-of-domain evaluation set that we construct.
These findings point to deep-rooted limitations in current benchmark design.

To address these shortcomings, we propose three essential directions for future multilingual MT benchmarks:
(1) source sentences should be less domain-specific and technical;
(2) benchmarks should minimize the influence of named entities in evaluation;
and (3) the content should avoid being overly centered on English-speaking cultural and linguistic norms.
These principles are crucial for the fair and meaningful MT evaluation across the world's languages.
\section{Related work}
A growing number of multilingual MT benchmarks have been proposed to assess the capabilities of modern MT systems across a wide range of languages.
Among them, the FLoRes series of benchmarks have become particularly influential in recent years.

The original Flores benchmark \cite{guzman-etal-2019-flores} was introduced to evaluate low-resource MT, specifically targeting Nepali--English and Sinhala--English translation.
It consisted of professionally translated Wikipedia sentences and aimed to highlight the performance gap between high- and low-resource languages.
Despite its narrow language scope, this work played an important role in initiating systematic evaluation for low-resource MT.

Flores-101 \citep{goyal-etal-2022-flores} extended this effort significantly by providing human-translated data for 3,001 English sentences in 101 languages, enabling over 10,000 translation directions.
The source sentences are curated to be multi-domain (WikiNews,\footnote{\url{https://en.wikinews.org}} WikiJunior,\footnote{\url{https://en.wikibooks.org/wiki/Wikijunior}} and WikiVoyage\footnote{\url{https://en.wikivoyage.org}}) and multi-topic (travel, politics, science, and crime, among others).
The benchmark follows a rigorous quality control process involving multiple rounds of professional translation and review.
It standardized multilingual MT evaluation by offering parallel test sets that are fully bitext-aligned, allowing direct comparison across language pairs.

FLORES-200 \citep{nllbteam2022languageleftbehindscaling} further expanded the benchmark to 200 languages, forming the backbone of the No Language Left Behind (NLLB) project \cite{nllb-scaling-neural-machine-translation-200}.
With over 40,000 translation directions, it represents the most comprehensive multilingual MT test suite to date.
The benchmark is currently managed and developed under the name of FLORES+,\footnote{\url{https://huggingface.co/datasets/openlanguagedata/flores_plus}} available under the CC-BY-SA-4.0 license.

Recent work has aimed to further improve the quality and coverage of FLORES benchmarks.
\citet{abdulmumin2024correctingfloresevaluationdataset} highlight systemic issues in the FLORES evaluation sets for four African languages, proposing corrections that enhance fluency and adequacy for more reliable MT evaluation.
In parallel, several efforts have sought to extend the benchmark to additional underrepresented languages.
For instance, \citet{gordeev-etal-2024-flores} introduce FLORES+ for Erzya, a severely low-resource Uralic language;
\citet{kuzhuget-etal-2024-enhancing} augment FLORES with a high-quality Tuvan test set;
\citet{perez-ortiz-etal-2024-expanding} expand coverage for Iberian languages;
\citet{ali-etal-2024-expanding} contribute new evaluation data for Emakhuwa, a Bantu language spoken in Mozambique; 
and \citet{yankovskaya-etal-2023-machine} focus on Finno--Ugric languages, addressing both translation quality and data availability.
\section{Human re-evaluation}

To investigate the translation quality of the FLORES+ benchmark, we manually re-evaluated translations of the four genealogically, orthographically, and geographically diverse languages with varying resource availability: Asante Twi, Japanese, Jinghpaw, and South Azerbaijani.
See Table~\ref{tab:languages} for details of the target languages.

\subsection{Setup}
For each language, we recruited a native speaker who is also fluent in English to serve as an annotator.
All of the annotators have had higher education in English and are currently affiliated with a university.
Their academic backgrounds are linguistics (Japanese and South Azerbaijani), education (Jinghpaw), and biological sciences (Asante Twi).
They have also had either professional or academic experience of translation involving their target language.
Each annotator is given pairs of English and the target language from the development set of the FLORES+ dataset corresponding to their language.
Their task is to assess the adequacy and fluency of each translation by comparing the English source sentence with its translation in the target language.
As we proceeded with the assessment, the manual evaluation and correction turned out to be time-consuming for the annotators due to the highly technical nature of the source sentences.
For this reason, we restrict the examined sentences to be the first 50 sentences of the dev set for this study.
The dataset version examined in this study is version 2.0 released on November 16, 2024.

\begin{table*}[t]
    \centering
    \setlength{\tabcolsep}{4pt}
    \small
    \begin{tabular}{lllll} \toprule
        Language & Code & Language family & Writing system & Region \\ \midrule
        Asante Twi & \texttt{twi\_Latn\_asan1239} & Akan $<$ Kwa $<$ Atlantic--Congo & Latin & West Africa \\
        Japanese & \texttt{jpn\_Jpan} & Japonic & Hiragana, Katakana, Kanji & East Asia \\
        Jinghpaw & \texttt{kac\_Latn} & Tibeto--Burman $<$ Sino--Tibetan & Latin & Southeast Asia \\
        South Azerbaijani & \texttt{azb\_Arab} & Oghuz $<$ Turkic & Perso--Arabic & West Asia \\
        \bottomrule
    \end{tabular}
    \caption{A list of the target languages and their characteristics.
    \emph{Code} refers to the corresponding language code in the FLORES+ dataset.
    Note that Jinghpaw is also referred to as ``Jingpho'', ``Jingpo'', or ``Kachin'' in different sources.
    }
    \label{tab:languages}
\end{table*}

\paragraph{Annotation guidelines.}
As per the original assessment guidelines in Flores-101 \cite{goyal-etal-2022-flores}, the annotators are instructed to categorize each sentence pair using the following labels:
    \emph{Correct},
    \emph{Wrong grammar},
    \emph{Wrong punctuation},
    \emph{Wrong spelling},
    \emph{Wrong capitalization},
    \emph{Inaccurate addition}, 
    \emph{Inaccurate omission},
    \emph{Mistranslation},
    \emph{Unnatural translation},
    \emph{Untranslated text},
    and \emph{Wrong register}.
Multiple labels can be selected per sentence to capture overlapping or compound issues.
When the annotators identify at least one error in a sentence, they are asked to categorize the severity of the error(s) into Critical errors, Major errors, and Minor errors.
The criteria for the error severity categorization also follow the guidelines in the original work \cite{goyal-etal-2022-flores}.
In addition, annotators are optionally invited to provide their own translation as well as free-form comments explaining their judgments or highlighting specific concerns about the translation.
This manual evaluation setup aims to identify not only obvious translation errors but also subtler issues such as fluency, naturalness, and contextual appropriateness, which automatic metrics often fail to capture.

\paragraph{Metric.}
The original work \cite{goyal-etal-2022-flores} uses a metric called \emph{Translation Quality Score} (TQS) to evaluate the translation quality; however, the detailed definition of this metric is not explained in their work.
For this reason, we tentatively define TQS as follows:
$$
\text{TQS} = \dfrac{3 \cdot C + 2\cdot E_m + 1\cdot E_M + 0 \cdot E_c}{3 \cdot (C + E_m  + E_M + E_c)}
$$
where $C$ is the number of correct sentences, $E_m$ the number of sentences identified as Minor errors, $E_M$ the number of sentences identified as Major errors, and $E_c$ the number of sentences identified as Critical errors.
We also employ a modified version of Multidimensional Quality Metrics (MQM) \cite{burchardt-2013-multidimensional}:
$$
\text{TQS}_\text{MQM} = 1 - \dfrac{e_m + 5 \cdot e_M + 10 \cdot e_c}{W},
$$
where $e_m$ is the number of Minor errors, $e_M$ the number of Major errors, $e_c$ the number of Critical errors, and $W$ the total word count.
$\text{TQS}_\text{MQM}$ generally yields higher scores than $\text{TQS}$, as it normalizes over word count, reducing the impact of errors.

\subsection{Results}

\begin{table}[ht]
    \centering
    \small
    \setlength{\tabcolsep}{4pt}
    \begin{tabular}{lrrrr}
    \toprule
    Category & \texttt{twi} & \texttt{jpn} & \texttt{kac} & \texttt{azb} \\
    \midrule
    Correct & 27 & 34 & 1 & 12 \\
    Wrong grammar & 4 & 0 & 13 & 3 \\
    Wrong punctuation & 1 & 0 & 0 & 6 \\
    Wrong spelling & 5 & 7 & 10 & 29 \\
    Wrong capitalization & 0 & 0 & 1 & 0 \\
    Inaccurately added information & 4 & 1 & 0 & 0 \\
    Inaccurately omitted information & 3 & 2 & 20 & 5 \\
    Mistranslation & 6 & 4 & 22 & 4 \\
    Unnatural translation & 8 & 4 & 22 & 3 \\
    Untranslated text & 1 & 25 & 9 & 1 \\
    Wrong register & 1 & 0 & 1 & 0 \\
    Other & 0 & 1 & 3 & 3 \\
    \bottomrule
    \end{tabular}
    \caption{Assessment statistics.
    The language codes correspond to the first three letters of the codes in Table~\ref{tab:languages}.
    Multiple category selection is allowed for sentences that are evaluated not to be correct.}
    \label{tab:error-categories}
\end{table}

\newcommand{\radius}[0]{1.5}
\newcommand{\correctcolor}[0]{green!40}
\newcommand{\minorcolor}[0]{yellow!40}
\newcommand{\majorcolor}[0]{orange!50}
\newcommand{\criticalcolor}[0]{red!50}
\newcommand{\legendfontsize}[0]{8pt}

\begingroup
\small
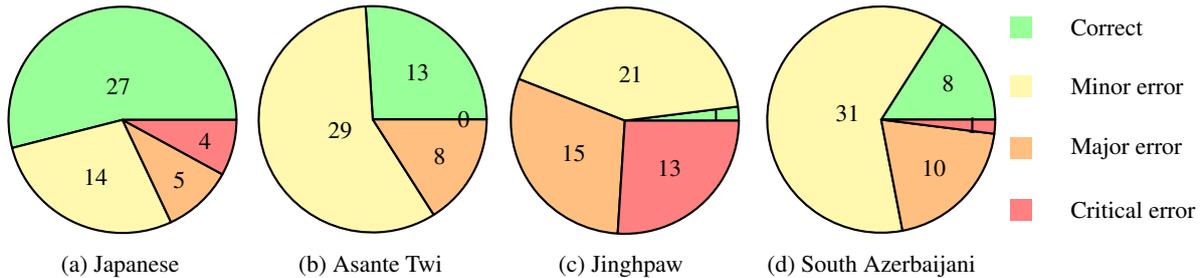
\begin{figure*}[t]
    \centering

    \begin{subfigure}[t]{0.2\textwidth}
        \centering
        \begin{tikzpicture}[font=\small]
            \pie[
                text={},
                radius=\radius,
                sum=auto,
                color={\correctcolor, \minorcolor, \majorcolor, \criticalcolor}
            ]{27/Correct, 14/Minor error, 5/Major error, 4/Critical error};
        \end{tikzpicture}
        \caption{Japanese}
    \end{subfigure}
    \begin{subfigure}[t]{0.2\textwidth}
        \centering
        \begin{tikzpicture}[font=\small]
            \pie[
                text={},
                radius=\radius,
                sum=auto,
                color={\correctcolor, \minorcolor, \majorcolor, \criticalcolor}
            ]{13/Correct, 29/Minor error, 8/Major error, 0/Critical error};
        \end{tikzpicture}
        \caption{Asante Twi}
    \end{subfigure}
    \begin{subfigure}[t]{0.2\textwidth}
        \centering
        \begin{tikzpicture}[font=\small]
            \pie[
                text={},
                radius=\radius,
                sum=auto,
                color={\correctcolor, \minorcolor, \majorcolor, \criticalcolor}
            ]{1/Correct, 21/Minor error, 15/Major error, 13/Critical error};
        \end{tikzpicture}
        \caption{Jinghpaw}
    \end{subfigure}
    \begin{subfigure}[t]{0.2\textwidth}
        \centering
        \begin{tikzpicture}[font=\small]
            \pie[
                text={},
                radius=\radius,
                sum=auto,
                color={\correctcolor, \minorcolor, \majorcolor, \criticalcolor}
            ]{8/Correct, 31/Minor error, 10/Major error, 1/Critical error};
        \end{tikzpicture}
        \caption{South Azerbaijani}
    \end{subfigure}
    \begin{tikzpicture}
        \matrix[
            column sep=1em,
            row sep=0.8em,
            nodes={anchor=west}
        ] {
            \node[fill=\correctcolor,
                minimum size=\legendfontsize,
                inner sep=0pt,
                shape=rectangle] {}; & \node{\footnotesize Correct}; \\
            \node[fill=\minorcolor,
                minimum size=\legendfontsize,
                inner sep=0pt,
                shape=rectangle] {}; & \node{\footnotesize Minor error}; \\
            \node[fill=\majorcolor,
                minimum size=\legendfontsize,
                inner sep=0pt,
                shape=rectangle] {}; & \node{\footnotesize Major error}; \\
            \node[fill=\criticalcolor,
                minimum size=\legendfontsize,
                inner sep=0pt,
                shape=rectangle] {}; & \node{\footnotesize Critical error}; \\
        };
    \end{tikzpicture}
    \caption{Judgement distribution by error type for the four languages.
    Each pie chart shows the number of Correct, Minor error, Major error, and Critical error annotations.}
    \label{fig:judgement_distributions}
\end{figure*}
\endgroup

Table~\ref{tab:error-categories} summarizes the results of the human re-evaluation.
Surprisingly, there was only one sentence in Jinghpaw that was evaluated to be correct.
Figure~\ref{fig:judgement_distributions} illustrates the error severities across the target languages.
In terms of the error severities, while Asante Twi contains no critical errors, 4 critical errors were found in Japanese, 13 in Jinghpaw, and 1 in South Azerbaijani.
Critical errors are ``issues that render the content unfit for use'' due to the original meaning seriously distorted ``in such a way that it becomes incomprehensible or that the essence of the message is lost'' \cite{goyal-etal-2022-flores}.
Table~\ref{tab:critical-errors} shows examples of the sentences identified as containing critical errors in Japanese and Jinghpaw.
The Japanese sentence contains a number of errors such as \emph{Inaccurate addition of information}, \emph{Grammatical error}, \emph{Untranslated text}, and \emph{Mistranslation}, leading to the loss of the original meaning.
The Jinghpaw translation contains a critical amount of \emph{Inaccurate omission of information} and overall \emph{Mistranslation} to the extent that it becomes incomprehensible.

\begin{table*}[t]
    \centering
    \small
    \newcommand{\trcolwidth}[0]{9.8em}
    \newcommand{\backtrcolwidth}[0]{14.4em}
    \newcommand{\correctcolwidth}[0]{12.3em}
    \newcommand{\error}[1]{\textcolor{red}{\hl{#1}}}
    \newcommand{\tagsize}[1]{{\tiny #1}}

     \begin{tabular}{@{}p{\trcolwidth}p{\trcolwidth}p{\backtrcolwidth}p{\correctcolwidth}@{}} \toprule
        English & Original & Backtranslation & Corrected \\ \midrule
       Hsieh also argued that the photogenic Ma was more style than substance. & (Japanese) \ja{謝長廷は、馬英九について実物よりも写真写りが良いと主張していました。} & Hsieh \error{Chang-ting}\tagsize{[Addition]} \error{was}\tagsize{[Grammar]} \error{also}\tagsize{[Untranslated]} \error{arguing}\tagsize{[Grammar]} that Ma \error{Ying-jeou}\tagsize{[Addition]} was \error{more photogenic than the actual appearance}\tagsize{[Mistranslation]}. & \ja{シエ氏はまた、写真写りの良いマー氏は、見た目ばかりで中身が伴っていないとも主張しました。} \\ \midrule
        At 11:20, the police asked the protesters to move back on to the pavement, stating that they needed to balance the right to protest with the traffic building up. & (Jinghpaw) Aten 11:20 hta dai masha ni hpe lam makau de sit na matu tsun wa ai, dai lam pat ai nta ni hpe atsawm lajang na matu ngu shana wa ai. & At 11:20, \error{they told the people to move sideways, they informed them to organize the houses that blocked the way}\tagsize{[Omission, Mistranslation]}. & Aten 11:20 hta, pyada ni hku na, dai ninghkap masha ni hpe lam makau de sit na matu tsun mat let, ninghkap na matu ahkaw ahkang nga ai hte maren, hkawm sa hkawm wa lam pat wa ai lam hpe mung, joi jang ra na matu shana wa ai. \\
        \bottomrule
    \end{tabular}
    \caption{Examples of critical errors found in the Japanese and Jinghpaw sets.
    The English source sentences are in the \emph{English} column, the original translations in the \emph{Original} column.
    For readers' convenience, the backtranslation from the original translation into English provided by the same annotators is added in the \emph{Backtranslation} column.
    The identified errors are highlighted with a red color with the error tags in square brackets.
    The \emph{Corrected} column provides the reference translation corrected by the annotators.
    }
    \label{tab:critical-errors}
\end{table*}

Table~\ref{tab:pilot-cer} summarizes the TQS and TQS$_\text{MQM}$ computed from the re-evaluation, as well as the character error rate (CER) and translation edit rate (TER) between the gold translation sentences in FLORES+ and the sentences corrected by the annotators.
TQS was substantially below the 90\% threshold for all the languages we examined.
For TQS$_\text{MQM}$, only Asante Twi exceeded 90\%, while the other three languages remained below the threshold.
In the worst cases, the Jinghpaw set had 76\% TER and the South Azerbaijani set 64\% TER.

\begin{table}[t]
    \centering
    \setlength{\tabcolsep}{3pt}
    \sisetup{detect-all, round-mode=places, round-precision=2}
    \begin{tabular}{@{}lSSSS@{}} \toprule
        Language & {CER} & {TER} & {TQS} & {TQS$_\text{MQM}$} \\ \midrule
        Jinghpaw & 44.547087503207594 & 76.17306520414382 & 40.0 & 70.89337175792507 \\
        Asante Twi & 15.529967522881605 & 26.041666666666668 & 70.0 & 94.0744920993228 \\
        Japanese & 13.225371120107962 & 2.901915264074289 & 76.0 & 83.37662337662337 \\
        South Azerbaijani & 16.421916942997  & 63.62672322375398 & 64.0 & 85.82995951417004 \\
        \bottomrule
    \end{tabular}
    \caption{CER and TER between the reference sentences and the sentences corrected by the assessors, as well as TQS and TQS$_\text{MQM}$.
    All values are in percent.
    Japanese sentences are tokenized by MeCab \cite{mecab}.}
    \label{tab:pilot-cer}
\end{table}

\subsection{Feedback from the annotators}

In addition to the quantitative evaluation, we collected qualitative feedback from human re-evaluators of the translated sentences.
Several recurring issues were observed in the evaluation across the target languages.

\paragraph{Named entities.}
The annotators commonly identified errors and unnaturalness regarding the translation and transliteration of NEs.
For example, ``Aldwych'' (a toponym in London), pronounced as /{\textprimstress}{\textopeno}{\textlengthmark}ldw{\textipa{I}}t{\textesh}/, is transliterated as \begin{CJK}{UTF8}{min}アルドウィック\end{CJK} /arudowikku/ in the benchmark translation, whereas it is commonly transliterated as \begin{CJK}{UTF8}{min}オールドウィッチ\end{CJK} /o{\textlengthmark}rudowitt{\textctc}i/.\footnote{
The strings enclosed in slashes represent the phonemic transcriptions.
The South Azerbaijani phonemic transcriptions are based on the Latin orthography.
}
In addition, when translating proper nouns made of English common nouns, the annotators were often confused whether they should just transliterate them or translate into the target language literally.
For example, ``Black Beauty'' (a name of a racing car) in the South Azerbaijani data was originally translated as
% \setfarsi\RL{قره گؤزل}
/qara göz{\textschwa}l/ in native words, while the annotator preferred the transliteration
% \setfarsi\RL{بلک بیوتی}
/bl{\textschwa}k byuti/.
How these NEs should be translated ultimately depends on the domain-specific knowledge about the entities, rather than the translator's grammatical competence.

\paragraph{Orthographic issues.}
The annotator for South Azerbaijani reported consistent spelling errors in the use of a zero-width non-joiner in the dataset.
In South Azerbaijani's Perso--Arabic orthography, the open front vowel /{\textschwa}/ in an open non-first syllable is written with \RL{h} or \RL{-h}, which also represents a voiceless glottal fricative consonant /h/.
To distinguish // from /h/, \RL{h}/\RL{-h} for /{\textschwa}/ is always detached from the next character by inserting a zero-width non-joiner.
For example, \setfarsi\RL{روزنامه‌} /ruznam{\textschwa}/ `newspaper', when suffixed by a third person possessive \setfarsi\RL{-سی} /si/, should be combined by a zero-width non-joiner as \setfarsi\RL{روزنامه‌سی} /ruznam{\textschwa}si/, and should not be directly conjoined with the previous letter as \setfarsi\RL{روزنامهسی}.
The South Azerbaijani sentences in FLORES+ often lack a zero-width non-joiner in such cases, and instead either a space is used or no separation at all (\textit{i.e.}, the character is combined with the preceding letter).
Although orthographic variation exists in modern written South Azerbaijani, the absence of a zero-width non-joiner for /{\textschwa}/ was strongly dispreferred by the annotator, as it may cause confusion with /h/.

\paragraph{Domain-specific concepts.}
While FLORES+ is a multi-domain dataset covering wide variety of topics, the annotators often expressed concerns that some sentences are too domain-specific involving specialized vocabulary from various fields.
These terms were sometimes unfamiliar even to fluent speakers.
For example, several sport-related concepts caused a confusion among the annotators.
The annotator for Jinghpaw noted that there is no widely accepted term in Jinghpaw for ``net point'' used in tennis, ``goal'' in soccer, and ``pit stop'' in motorsport.
The annotators for Asante Twi and Jinghpaw both noted that they did not know how to translate ``season'' in the context of sports.
The annotator for Japanese reported not knowing the term ``co-driver'' in the context of auto racing.

\paragraph{Grammar-specific issues.}
Japanese has honorific registers that are used when the speaker conveys politeness, respect, or humbleness \cite{harada-1976-honorifics}.
From its nature, the politeness register typically presupposes a hearer and is often used in spoken settings such as conversations and TV news.
In contrast, the neutral register is primarily used in a written form, such as books and newspaper.
The Japanese translations in FLORES+ were consistently rendered in the polite register, despite the fact that the sources of the sentences are from written media such as WikiNews.
In addition, the annotators for Asante Twi and Jinghpaw noted that some long English sentences containing multiple clauses with syntactic coordinations and relativizations were difficult to translate directly in one sentence.
To maintain fluency, they often had to split these into multiple sentences.

\paragraph{Cultural issues.}
For Jinghpaw, fundamental lexical gaps from English were evident.
The annotator for Jinghpaw noted the absence of native equivalents for certain concepts such as ``spring'' (as in four seasons) in the language because the region where Jinghpaw is spoken (Kachin State, Myanmar) falls into a (sub)tropical climate that does not have four seasons.

These limitations underscore the difficulty of translating domain-specific, technical English sentences into languages with different grammatical, lexical and conceptual systems and different cultural backgrounds.
It is also important to highlight that these systematic issues were not only observed in low-resource languages but also in a high-resource language like Japanese.
\section{Experiments}
In this section, we conduct two experiments to address the issues of the benchmark.
Based on the findings in the human re-evaluation that the sentences frequently contain NEs, the first experiment tests whether synthetically generated translations made of NEs copied from the English source sentences yield non-zero evaluation scores.
In the second experiment, we evaluate MT models with FLORES+ and our in-house evaluation datasets and examine the implications of using the benchmark for evaluating low-resource MT models.
The FLORES+ version used in the experiments is version 2.0 released on November 16, 2024.
The in-house evaluation datasets will be made publicly available.

\subsection{Named-entity copying}\label{sec:ne-copying}
The objective of this experiment is to assess the extent to which NEs contribute to the evaluation scores in the FLORES+ benchmark.
Our motivation stems from the observation that if the evaluation sentences contain a significant number of NEs, MT systems that merely copy those entities could achieve artificially inflated scores, thereby undermining the benchmark's ability to reflect true translation quality.

\subsubsection{Setup}

To simulate this scenario, we constructed a set of synthetic translations for each target language.
For each source sentence, we extracted the NEs and used them as the ``translated'' output.
To bypass the brevity penalty in BLEU (see also Appendix~\ref{sec:bleu}), we appended the string `` \texttt{dummy}'' 50 times after the extracted NEs.
We used gpt-4o to identify the named entities.
We then computed the BLEU \cite{papineni-etal-2002-bleu} and ChrF++ \cite{popovic-2015-chrf} scores of these synthetic translations against the official FLORES+ references in the languages with Latin-based orthographies.

In our experiments, we consider two scenarios of the state of hypothesis strings: (1) the hypothesis string is empty (\textit{i.e.}, no NE is detected in the source sentence), and (2) the hypothesis string consists of at least one matching word (\textit{i.e.}, NE).
In case of (1), the BLEU and ChrF++ scores of the synthetic sentence is 0, while (2) yields BLEU and ChrF++ scores larger than 0.
Under the assumption that the core objective of a translation task is not to just copy certain words but to test the ability of mapping the unique grammar and lexicon from one language to the other, the BLEU scores of those synthetic sentences in an multilingual MT evaluation benchmark should ideally be 0.

\subsubsection{Results}

\begin{figure*}[t]
    \centering
    \includegraphics[width=1\linewidth]{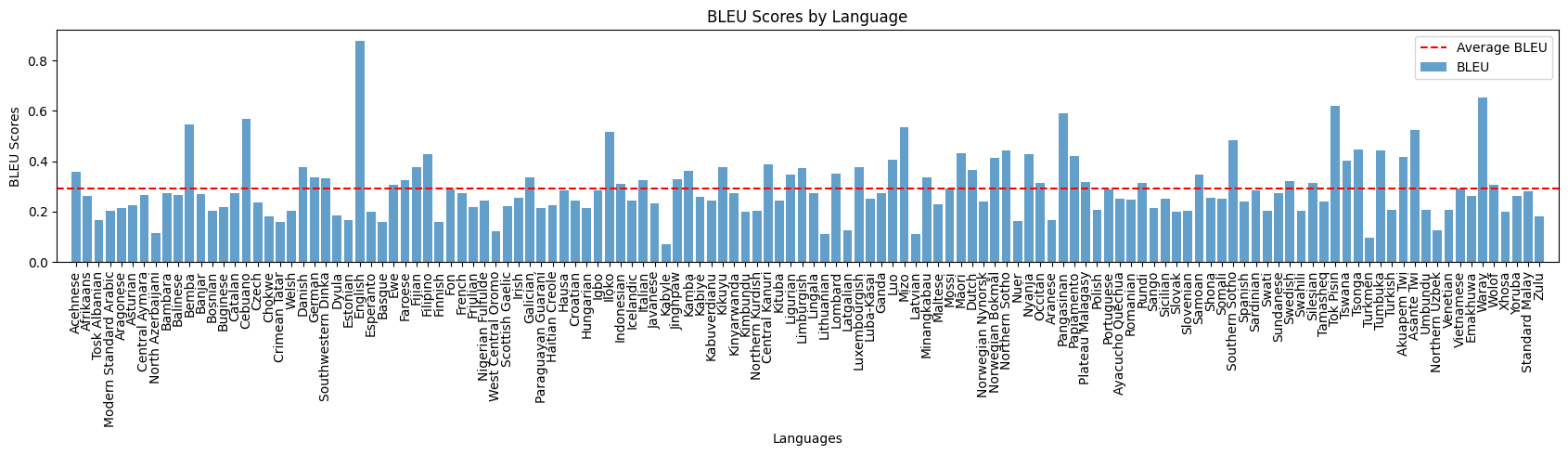}
    \includegraphics[width=1\linewidth]{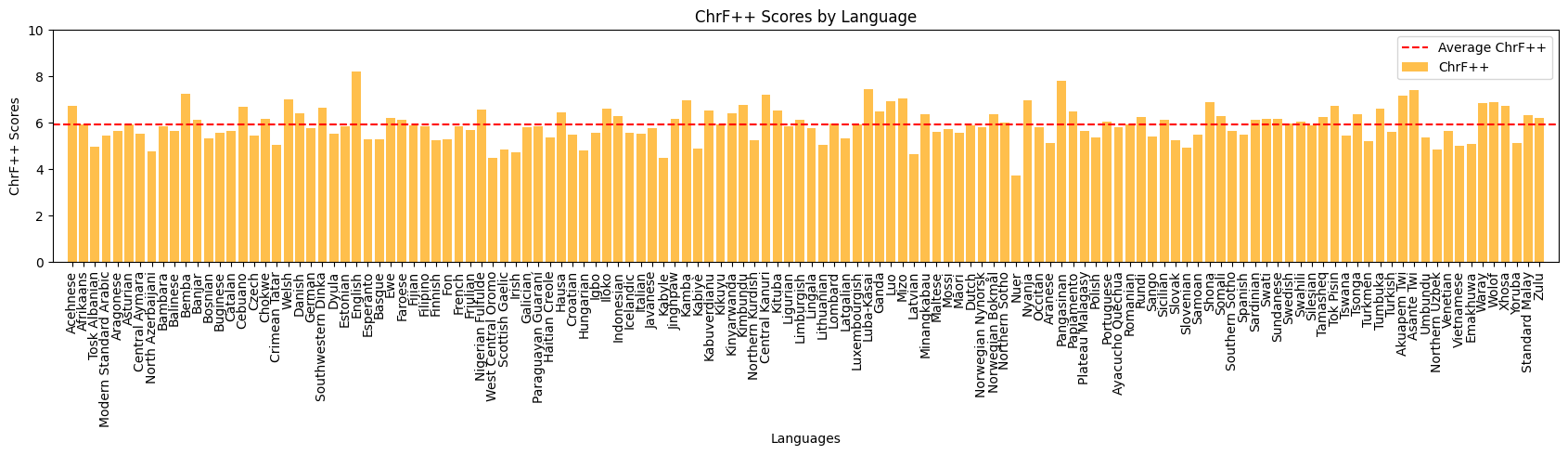}
    \caption{BLEU scores (top) and ChrF++ scores (bottom) of dummy translations with named entities copied from the source sentence.
    The tested languages are those with a Latin-based orthography.}
    \label{fig:flores_ner}
\end{figure*}

As the results in Figure~\ref{fig:flores_ner} demonstrate, even this trivial NE-copying baseline achieved non-zero scores across all the languages, indicating that the benchmark may reward surface-level lexical overlap rather than meaningful translation quality.
These results call into question the adequacy of current evaluation practices in multilingual MT with FLORES+ because MT systems that are good at copying NEs yield a better score than those that are not.

\subsection{Jinghpaw MT}
As the second experiment, we train an MT model for Jinghpaw--English translation based on the pretrained NLLB models \cite{nllbteam2022languageleftbehindscaling} and see the implication of using the benchmark.

\subsubsection{Setup}

\paragraph{Datasets.}
\begin{table}[t]
    \centering
    \small
    \begin{tabular}{lrrrr} \toprule
         & {Train} & {Dev} & {Test} & {Total} \\ \midrule
        PARADISEC (P) & 38113 & 755 & 743 & 39611 \\
        Dictionary (D) & 3151 & 0 & 0 & 3151 \\
        Dialogue & 0 & 0 & 222 & 222 \\
        NLLB (N) & 121081 & 0 & 0 & 121081 \\
        FLORES+ & 0 & 997 & 1011 & 2008 \\
        \midrule
        Total & 162345 & 1752 & 1976 & 166073 \\
        \bottomrule
    \end{tabular}
    \caption{The number of parallel sentences contained in the datasets used in this experiment.
    Note that we use the \emph{Devtest} set of FLORES+ as the test set.}
    \label{tab:datasets}
\end{table}

The datasets used in the experiments include Jinghpaw narrative data from PARADISEC \cite{kurabe-2013-kachin-folktales, kurabe-2017-kachin-culture-history}, example sentences listed in the \textit{Jinghpaw Usage Dictionary} \cite{kurabe-2020-jinghpaw-dictionary}, dialogues from the \textit{Jinghpaw Reader} \cite{kurabe-2020-jinghpaw-reader}, the Jinghpaw--English parallel sentences from the NLLB dataset,\footnote{\url{https://opus.nlpl.eu/NLLB/corpus/version/NLLB}} and the FLORES+ benchmark; see Table~\ref{tab:datasets} for details.
Among these, due to the small number of samples, all the Dictionary sentences are used in the training data, and all the Dialogue sentences are used in the test data.
Additionally, because the quality of the NLLB dataset is not guaranteed, we remove duplicate sentence pairs and sentence pairs with significantly mismatched lengths between the two languages based on the method used by \citet{deguchi-etal-2023-naist} and use the remaining data in the training data.
The NLLB dataset originally contained 1,003,100 sentence pairs, which was reduced to 121,081 after filtering.
Since FLORES+ is intended for evaluation, we use the dev set as validation data and the devtest set as test data and do not use it at all for training.

\paragraph{Model training.}

In this experiment, we set the baselines as the pretrained multilingual MT models NLLB-600M\footnote{\url{https://huggingface.co/facebook/nllb-200-distilled-600M}} and NLLB-1.3B\footnote{\url{https://huggingface.co/facebook/nllb-200-1.3B}} and fine-tune them for Jinghpaw$\leftrightarrow$English MT using the different sets of training data described in Table~\ref{tab:datasets}.
We perform fine-tuning separately for both translation directions: from Jinghpaw to English and from English to Jinghpaw.
The training objective is to minimize the cross-entropy loss, as in the original NLLB models.

For all model training runs, the batch size is set to 8, the learning rate to $0.0001$, the number of warm-up steps to 500, the learning rate decay factor per epoch to 0.9, and the maximum number of epochs to 20.
Other hyperparameters follow the default settings of the NLLB models.
All model fine-tuning is conducted using a single A10 GPU.

For evaluation, we employ BLEU and ChrF++ as the evaluation metrics.
After the training, the checkpoint with the highest BLEU score on the validation set is picked for evaluation.
As the evaluation datasets, we use the test set of PARADISEC, the devtest set of FLORES+, and the Dialogue dataset.

\subsubsection{Results}

Results in Table~\ref{tab:nllb-mt} present the BLEU and ChrF++ scores for both translation directions: Jinghpaw$\rightarrow$English and English$\rightarrow$Jinghpaw.
We compare the pretrained NLLB models (600M and 1.3B) against their fine-tuned variants using various subsets of our collected datasets: the Dictionary dataset (D), the PARADISEC dataset (P), and the filtered NLLB dataset (N).

Across both translation directions, we observe that the pretrained NLLB models achieve the highest scores on the FLORES+ set, particularly for the larger NLLB-1.3B model.
However, their performance degrades substantially on more natural and simpler sentences from the PARADISEC and Dialogue test sets.
For example, in the Jinghpaw$\rightarrow$English setting, the NLLB-1.3B baseline achieves a BLEU score of only 2.29 on PARADISEC data, despite scoring 13.95 on FLORES+.
This suggests that these models are tuned for data that share domains similar to FLORES+.

\begin{table*}[!t]
    \centering
    \small
    \begin{subtable}[t]{0.95\textwidth}
        \centering
        \begin{tabular}{llcccc} \toprule
            Model & Training data & PARADISEC (test) & FLORES (devtest) & Dialogue & Average \\ \midrule
            \multirow{5}{6.5em}{NLLB-600M} & Baseline & 2.32 / 20.34 & \textbf{12.77} / \textbf{35.47} & 17.35 / 32.42 & 10.81 / 29.41 \\
            & D & 4.58 / 22.52 & 9.73 / 31.44 & 18.08 / 34.43 & 10.80 / 29.46 \\
            & P & \textbf{12.58} / 31.35 & 5.91 / 27.34 & 22.41 / 39.78 & 13.63 / 32.82 \\
             & P+D & 12.55 / \textbf{31.91} & 6.04 / 27.12 & 22.30 / 40.38 & 13.63 / 33.14 \\
            & P+D+N & 12.86 / 31.91 & 6.88 / 30.03 & \textbf{23.35} / \textbf{41.67} & \textbf{14.36} / \textbf{34.54} \\
            \midrule
            \multirow{5}{7em}{NLLB-1.3B} & Baseline & 2.29 / 19.72 & \textbf{13.95} / \textbf{37.27} & 16.66 / 33.44 & 10.97 / 30.14 \\
            & D & 4.23 / 20.44 & 10.90 / 30.56 & \textbf{24.57} / \textbf{41.71} & 13.23 / 30.90 \\
            & P & 12.95 / 32.11 & 5.48 / 26.59 & 18.09 / 34.84 & 12.17 / 31.18 \\
            & P+D & 12.95 / 31.45 & 5.41 / 26.66 & 23.49 / 40.96 & \textbf{13.95} / \textbf{33.02} \\
            & P+D+N & \textbf{13.40} / \textbf{32.19} & 4.87 / 27.25 & 22.44 / 39.81 & 13.57 / 33.08 \\
            \bottomrule
        \end{tabular}
        \caption{Jinghpaw$\rightarrow$English MT.}
        \label{tab:nllb-kaceng}
    \end{subtable}
    
    \vspace{0.8em}
    
    \begin{subtable}[t]{0.95\textwidth}
        \centering
        \begin{tabular}{llcccc} \toprule
            Model & Training data & PARADISEC (test) & FLORES (devtest) & Dialogue & Average \\ \midrule
            \multirow{5}{6em}{NLLB-600M} & Baseline & 3.67 / 25.32 & \textbf{9.68} / \textbf{34.43} & 13.05 / 39.61 & 8.80 / 33.12 \\
            & D & 4.61 / 26.77 & 8.06 / 31.84 & 17.10 / 41.36 & 9.92 / 33.32 \\
            & P & 10.99 / 32.92 & 1.65 / 21.76 & 18.90 / 44.28 & 10.51 / 32.99 \\
            & P+D & 11.40 / 33.87 & 2.50 / 24.17 & \textbf{21.75} / 44.82 & 11.88 / 34.29 \\
            & P+D+N & \textbf{11.43} / \textbf{34.35} & 3.94 / 24.02 & 21.17 / \textbf{45.68} & \textbf{12.18} / \textbf{34.68} \\
            \midrule
            \multirow{5}{7em}{NLLB-1.3B} & Baseline & 4.15 / 25.94 & \textbf{10.97} / \textbf{36.00} & 16.95 / 43.02 & 10.69 / 34.99 \\
            & D & 4.48 / 26.05 & 6.12 / 29.68 & 20.17 / 44.34 & 10.26 / 33.36 \\
            & P & 12.19 / 35.04 & 2.26 / 24.09 & \textbf{24.41} / \textbf{48.21} & 12.95 / 35.78 \\
            & P+D & 12.30 / 35.75 & 3.74 / 27.45 & 22.75 / 48.01 & 12.93 / \textbf{37.07} \\
            & P+D+N & \textbf{12.78} / \textbf{35.87} & 3.60 / 25.16 & 23.01 / 47.82 & \textbf{13.13} / 36.28 \\
            \bottomrule
        \end{tabular}
        \caption{English$\rightarrow$Jinghpaw MT.}
        \label{tab:nllb-engkac}
    \end{subtable}
    
    \caption{Results of English$\leftrightarrow$Jinghpaw MT.
    Each row corresponds to a training setting, and each column shows performance on a specific evaluation set.
    D stands for the Dictionary dataset, P for the PARADISEC dataset, and N for the NLLB dataset.
    The Baselines are the pre-trained NLLB models without fine-tuning.
    In each cell, the score on the left-hand side represents the BLEU score, and the one on the right-hand side the ChrF++ score.}
    \label{tab:nllb-mt}
\end{table*}

In contrast, fine-tuning on naturalistic datasets such as the PARADISEC corpus and the Dictionary examples substantially improves performance on the PARADISEC (test) data and the Dialogue data.
For instance, the NLLB-600M model fine-tuned on P+D+N reaches 23.35 BLEU on the Dialogue test set, outperforming the baseline by a wide margin.
A similar trend holds in the English$\rightarrow$Jinghpaw direction, where models fine-tuned on the naturalistic datasets consistently outperform the baseline on non-FLORES+ test sets.

Notably, fine-tuning on PARADISEC and Dictionary typically results in a drop in FLORES+ performance, suggesting a mismatch between their domains.
This performance trade-off raises a critical question: What does it mean to ``perform well'' on a benchmark in which the reference translations themselves are questionable?
In our human evaluation, for instance, only one out of the fifty Jinghpaw FLORES+ sentences was judged to be a correct translation by a native speaker.

The results of the two experiments in this section highlight two key insights:
(1) FLORES+ may reward surface-level alignment with domain-specific NEs rather than generalizable translation ability; 
and (2) fine-tuning on small but naturalistic corpora can significantly enhance model performance on real-world test sets, albeit hurting the benchmark scores.
These findings underscore the urgent need for more representative evaluation datasets that reflect the linguistic and cultural diversity of speaker communities.
\section{Concluding remarks}

In this study, we critically examined the FLORES+ multilingual MT benchmark through human re-evaluation and empirical analysis of four languages.
Our human re-evaluation revealed that the translation quality in these languages falls below the benchmark's reported standard, calling into question the reliability of its quality claims.
Annotators consistently reported that many English source sentences were highly domain-specific or culturally loaded, making them difficult to translate naturally or at all in some languages.
These issues were observed not only in the low-resource languages but also in a high-resource language like Japanese.
This raises fundamental concerns about the benchmark's linguistic inclusivity and representativeness.

Our experimental analyses further uncovered structural issues in the benchmark design.
In the named-entity copying experiment, mock translations that merely reproduced named entities from the source achieved non-trivial BLEU and ChrF++ scores.
This suggests that current evaluation protocols may reward superficial copying over true translational adequacy and fluency.
Moreover, in our Jinghpaw MT experiments, we found that while the pretrained NLLB models performed best on the FLORES+ set, they performed poorly on more naturalistic evaluation datasets constructed from narratives and conversational texts.
In contrast, models fine-tuned on authentic, community-sourced data demonstrated improved performance on these realistic test sets, though their FLORES+ scores declined.

Taken together, these findings highlight the misalignment between benchmark-centered evaluation and the realities of multilingual MT, particularly for low-resource and culturally diverse languages.
To address these limitations, we propose the following directions for future benchmark development:
(1) source sentences should be less domain-specific and more universally interpretable;
(2) the influence of NEs should be minimized to better assess true translation capabilities such as grammatical and lexical knowledge;
(3) content should avoid undue bias toward English-speaking cultural and linguistic norms;
and (4) ongoing efforts to revise and expand existing datasets should be actively supported and valued.
\section*{Limitations}
While our study reveals critical insights into the shortcomings of the FLORES benchmark and highlights challenges in multilingual MT evaluation, it is not without limitations.

First, our human re-evaluation was conducted on a relatively small subset of the benchmark (50 sentence pairs per language) and focused on a limited set of languages.
Although the languages were chosen to represent typological and resource-level diversity, further large-scale evaluations across more language families are necessary to generalize our findings.

Second, the re-evaluation study was conducted by one annotator per language due to the scarcity of qualified annotators for the languages examined.
We acknowledge that relying on a single annotator per language limits the objectivity of the study and makes it impossible to measure inter-annotator agreement.

Third, while we critique the limitations of FLORES+, we do not propose a fully realized alternative benchmark.
Future work should explore concrete methods for constructing domain-general, culturally inclusive evaluation sets that are robust to lexical and structural biases.

Last but not least, we fully acknowledge the substantial effort and coordination required to construct the original FLORES benchmark, particularly the translation of over 3,000 sentences into more than 200 languages.
This undertaking has significantly advanced multilingual MT research and helped bring attention to low-resource languages in the NLP community.
Our critiques are not meant to diminish this achievement, but rather to support its ongoing improvement.
We hope our findings contribute to the broader conversation around building more inclusive, culturally aware, and contextually appropriate evaluation benchmarks.
\section*{Ethical considerations}

This work aims to critically assess the limitations of multilingual MT benchmarks with a focus on underrepresented and low-resource languages.
All human evaluations in this study were conducted by native speakers of the target languages who were fluent in English.
Annotators were fully informed of the purpose of the study and participated voluntarily.
No sensitive or offensive content was collected during the study, and any information that could identify annotators has been anonymized.

Our findings underscore the ethical implications of benchmark design in NLP and MT.
Benchmarks that rely heavily on English-centric, domain-specific, or culturally narrow content can systematically disadvantage certain languages and communities.
When such benchmarks are treated as gold standards for model evaluation, they risk misrepresenting the capabilities of MT systems and, more seriously, may perpetuate linguistic inequities, leaving many languages behind.

We emphasize the importance of language technologies that respect speaker communities, their linguistic norms, and their cultural contexts.
Translation systems should not merely map words across languages and copy or transliterate proper nouns but should support meaningful, context-aware communication.
This requires training and evaluating models on culturally informed, naturally occurring data rather than relying solely on synthetically generated or automatically translated corpora, which may introduce artifacts or distort local usage, leading to unnatural sentences or translationese.

The datasets used in our study, including the PARADISEC narrative corpus, the Jinghpaw Dictionary, and the Jinghpaw Reader, are all publicly available and were developed with care and proper documentation.
We advocate for ongoing, collaborative partnerships with speaker communities in both data creation and model evaluation.
Ethical NLP research must prioritize not just linguistic diversity in principle, but the lived realities, needs, and voices of language users.

While we used AI assistants such as ChatGPT to support the writing process in a non-native language, all content was carefully reviewed and edited to ensure accuracy, clarity, and alignment with our research objectives.

\section*{Acknowledgments}
This material is based upon work supported by the National Science Foundation under grant BCS-2109709 and IIS-2137396 and by the Japan Society for the Promotion of Science under KAKENHI grant JP24K03887.

\bibliography{main}

\appendix
\section{Implementation details of metrics}
\label{sec:appendix}

\subsection{BLEU}\label{sec:bleu}
For computing the BLEU scores, we used the implementation provided by the sacrebleu library (version 2.5.1) \cite{post-2018-call}.
As given in the body of this paper, the general formula for calculating a BLEU score is:
$$
\text{BLEU} = \text{BP} \cdot \exp\left(\sum_{n=1}^N w_n\log p_n\right),
$$
where BP is the brevity penalty
$\text{BP} = \min(1, e^{1 - \frac{r}{c}})$,
$N$ is the maximum value of $n$ for counting $n$-grams,
$w_n$ the weight for each $n$,
$r$ the length (number of words, split by whitespaces) of the reference sentence,
$c$ the length of the hypothesis, and
$p_n$ the modified $n$-gram precision, which is the number of matches devided by the number of $n$-grams.
The number of matches of a word considered in the modified $n$-gram precision is at most the frequency of the word in the reference text.
In our experiments, we used the default hyperparameters implemented by sacrebleu: $N = 4$ and $w_n = \frac{1}{4}$.
To avoid the cases of $p_n = 0$, which is often the case in our experiments due to a number of empty hypothesis strings in the named-entity-copying dummy translations, we made sure that $p_n$ is smoothed by exponential decay when $p_n = 0$.
Exponential decay smoothing is the default setting in sacrebleu.

\begin{algorithm}[H]
\caption{Exponential Decay for BLEU}
\begin{algorithmic}[1]
\Require $n$-gram match counts: $\text{match}[1 \dots N]$
\Require Total candidate $n$-grams: $\text{total}[1 \dots N]$
\Ensure Smoothed $n$-gram precisions $p_1, \dots, p_N$

\State $s \gets 1$ \Comment{Initialize decay factor}
\For{$n \gets 1$ to $N$}
    \If{$\text{correct}[n] = 0$}
        \State $s \gets 2 \cdot s$ \Comment{Exponential penalty}
        \State $p_n \gets \frac{1}{s \cdot \text{total}[n]} \cdot 100$
    \Else
        \State $p_n \gets \frac{\text{correct}[n]}{\text{total}[n]} \cdot 100$
    \EndIf
\EndFor
\end{algorithmic}
\end{algorithm}

\subsection{ChrF++}

We also used sacrebleu for computing ChrF++ score.
The formula for computing ChrF++ score is:
$$
\text{ChrF++} = \dfrac{(1 + \beta^2) P \cdot R}{\beta^2 P + R},
$$
In the typical implementation of ChrF++, which we followed in this study, uses $\beta=2$, $N_w = 2$ and $N_c = 6$, where $N_w$ is the maximum $n$-gram considered at the word level and $N_c$ at the character level.
The sentence-level precision $P$ is the averaged value of character-level and word-level precisions, and the sentence-level recall $R$, similarly, is the averaged value of character-level and word-level recalls: 
$$
P = \dfrac{P_c + P_w}{N_c + N_w},
R = \dfrac{R_c + R_w}{N_c + N_w}.
$$

The corner cases where the reference or hypothesis text is shorter than $N_c$ or $N_w$ are ignored; that is, if we have $N_w \gets 2$ but the hypothesis length is $1$, then the $2$-gram is ignored and $N_w \gets 1$.
In our experiments, we removed whitespaces when computing $P_c$ and $R_c$, which is the default configuration in sacrebleu.
Also, we made sure that the ChrF++ calculation correctly considers word-level 2-grams by specifying \texttt{sacrebleu.CHRF(word\_order=2)}.

\section{Prompt template for named-entity extraction}
The prompt template used to extract NEs for generating dummy translation sentences in the experiment described in Section~\ref{sec:ne-copying} is provided below:

\begin{tcblisting}{listing only}
Please output the extract named entities concatenated by whitespace.

For example,
**Input:** On Sunday, May 4, in Peru's Pataz province in the northern Department of La Libertad region, near one of Peru's largest gold mines, police has found bodies of thirteen security guards who were kidnapped on Saturday, April 26, allegedly by individuals involved in illegal mining.
**Output:** Sunday May 4 Peru Pataz province Department of La Libertad region Peru Saturday April 26

{other_in_context_examples}

Now, it's your turn.
**Input:** {text}
**Output:**
\end{tcblisting}

\section{Sample sentences from the in-house datasets}

Table~\ref{tab:sample-sentences} shows some sample sentences from the in-house datasets used in this study.

\begin{table*}[h]
    \centering
    \begin{tabular}{lp{14em}p{14em}} \toprule
         & Jinghpaw & English \\ \midrule
        PARADISEC & Dai hka hpe rap na matu shan lahkawng gaw grai yak ai da. & It was really difficult for them to cross the river. \\
        Dictionary & ndai nye a hkawhkam nang shayi hpe nang hpe ap sana & I'll leave this princess of mine to you \\
        Dialogue & Manau poi hte seng ai laika buk ka na matu myit da ai. & I hope to write a book about the Manau festival. \\
        \bottomrule
    \end{tabular}
    \caption{Sample sentences from the in-house datasets used in this study.}
    \label{tab:sample-sentences}
\end{table*}

\section{Details of instructions given to the annotators}
The data collection process was conducted in consultation with the Institutional Review Board at the university which this study was primarily conducted at.
The instructions given to the annotators are provided below:

\small
\begin{tcblisting}{listing only}
## Instruction
1. Open the Spreadsheet {link}
2. For each row, compare the sentences in English and {language}, and judge whether the {language} sentence is a correct translation of the English sentence. Specify  your judgment(s) in the "Fine-grained evaluation" column. Specify your judgment(s) in the "Evaluation" column. See the guidelines in Goyal et al. (2021).
3. If the sentence contains an error(s), specify the severity of the error(s) in the "Error severity" column. See the guidelines in Goyal et al. (2021).
4. If you think that the translation has a problem (the evaluation is not "Correct"), please provide your correct translation in the "Corrected {language} translation" column.
5. If you have any other comments, such as any additional clarification, questions, etc., feel free to write them in the "Comments" column.
6. Repeat steps 2 - 5 for the 50 sentences (up to the 51st row in the spreadsheet)
\end{tcblisting}

\end{document}